\def\year{2022}\relax
\documentclass[letterpaper]{article} 
\usepackage{aaai22}  
\usepackage{times}  
\usepackage{helvet}  
\usepackage{courier}  
\usepackage[hyphens]{url}  
\usepackage{graphicx} 
\usepackage{natbib}  
\usepackage{caption} 
\usepackage{latexsym}
\usepackage{paralist}
\usepackage{amsmath}
\usepackage{amsfonts}
\usepackage{xcolor}
\usepackage{amssymb}
\usepackage{multicol}
\usepackage{multirow}
\usepackage{booktabs}
\usepackage{mathrsfs}
\usepackage{colortbl}
\usepackage{fdsymbol}
\definecolor{Gray}{gray}{0.8}
\definecolor{lGray}{gray}{.95}

\DeclareCaptionStyle{ruled}{labelfont=normalfont,labelsep=colon,strut=off} 
\usepackage{algorithm}
\usepackage{algorithmic}

\usepackage{newfloat}
\usepackage{listings}
\usepackage{pifont}

\newcommand{\cmark}{\ding{51}}
\newcommand{\xmark}{\ding{55}}

\floatstyle{ruled}
\newfloat{listing}{tb}{lst}{}
\floatname{listing}{Listing}
\title{{E}ntailment {R}elation {A}ware {P}araphrase {G}eneration}

\title{{E}ntailment {R}elation {A}ware {P}araphrase {G}eneration}
\author {
    Abhilasha Sancheti,\textsuperscript{\rm $\vardiamondsuit$ $\clubsuit$}
    Balaji Vasan Srinivasan, \textsuperscript{\rm $\clubsuit$}
    Rachel Rudinger \textsuperscript{\rm $\vardiamondsuit$}
}
\affiliations {
    \textsuperscript{\rm $\vardiamondsuit$} University of Maryland, College Park\\
    \textsuperscript{\rm $\clubsuit$} Adobe Research\\
    sancheti@umd.edu, rudinger@umd.edu, balsrini@adobe.com
}

\begin{document}

\maketitle
\begin{abstract}
We introduce a new task of \textit{entailment relation aware paraphrase generation} which aims at generating a paraphrase conforming to a given entailment relation ({\it e.g.} equivalent, forward entailing, or reverse entailing) with respect to a given input. We propose a reinforcement learning-based weakly-supervised paraphrasing system, \textbf{\texttt{ERAP}}, that can be trained using existing paraphrase and natural language inference (NLI) corpora without an explicit task-specific corpus.
A combination of automated and human evaluations show that \texttt{ERAP} generates paraphrases conforming to the specified entailment relation and are of good quality as compared to the baselines and uncontrolled paraphrasing systems. Using \texttt{ERAP} for augmenting training data for downstream textual entailment task improves performance over an uncontrolled paraphrasing system, and introduces fewer training artifacts, indicating the benefit of explicit control during paraphrasing.
\end{abstract}

\section{Introduction} \label{sec:intro}
Paraphrase is ``an alternative surface form in the same language expressing the same semantic content as the original form"~\citep{madnani2010generating}.
Although the logical definition of paraphrase requires strict semantic equivalence (or bi-directional entailment~\citep{androutsopoulos2010survey}) between a sequence and its paraphrase, data-driven paraphrasing
accepts a broader definition of \textit{approximate} semantic equivalence~\citep{bhagat2013paraphrase}.  Moreover, existing automatically curated paraphrase resources do not align with this logical definition. For instance, pivot-based paraphrasing rules extracted by~\citet{ganitkevitch2013ppdb}
contain hypernym or hyponym pairs, {\it e.g.} due to variation in the discourse structure of translations, and unrelated pairs, {\it e.g.} due to misalignments or polysemy in the foreign language. 
\begin{figure}[!t]
\scriptsize
    \centering
    \includegraphics[width=0.4\textwidth,height=5.2cm]{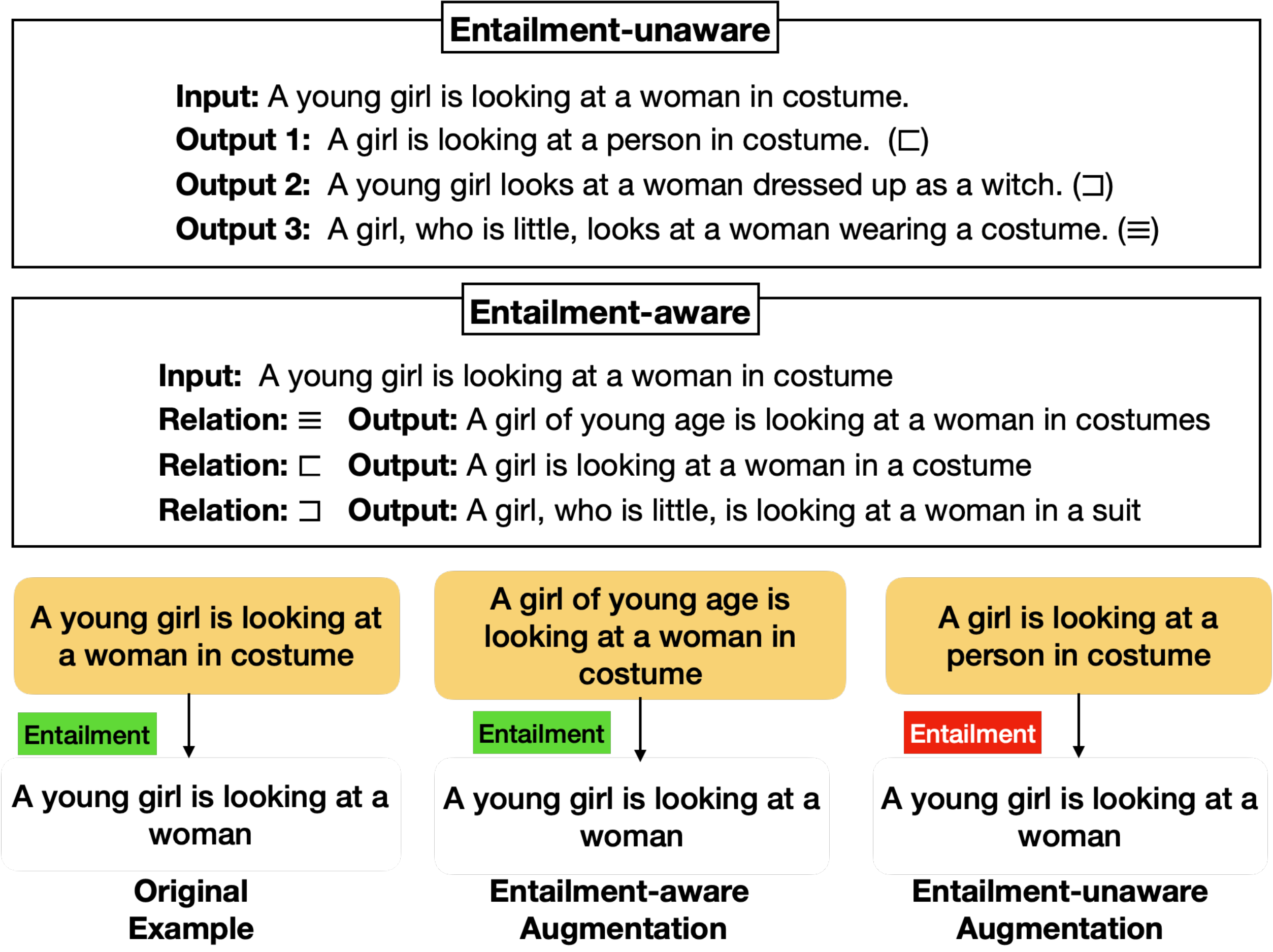}
    \caption{Entailment-unaware system might output \textit{approximately} equivalent paraphrases. Label preserving augmentations generated using such system for textual entailment task can result in incorrect labels (red). Explicit entailment relation control in entailment-aware system helps in reducing such incorrectly labeled augmentations (green).}
    \label{fig:task}
\end{figure}

While this flexibility of \textit{approximate} semantic equivalence allows for greater diversity in expressing a sequence, it comes at the cost of the ability to precisely control the semantic entailment relationship (henceforth ``entailment relation") between a sequence and its paraphrase. This trade-off severely limits the applicability of paraphrasing systems or resources to a variety of downstream natural language understanding (NLU) tasks ({\it e.g.} machine translation, question answering, information retrieval, and natural language inferencing~\citep{pavlick2015adding}) (Figure~\ref{fig:task}).
For instance, semantic divergences in machine translation have been shown to degrade the translation performance~\citep{carpuat2017detecting,pham2018fixing}. 

Existing works identify directionality (forward, reverse, bi-directional, or no implication) of paraphrase and inference rules~\citep{bhagat2007ledir}, and add semantics (natural logic entailment relationships such as equivalence, forward or reverse entailment, etc.) to data-driven paraphrasing resources~\citep{pavlick2015adding} leading to improvements in lexical expansion and proof-based RTE systems, respectively. However, entailment relation control in paraphrase generation is, to our knowledge, a relatively unexplored topic, despite its potential benefit to downstream applications~\citep{madnani2010generating} such as Multi-Document Summarization (MDS) (or Information Retrieval (IR)) wherein 
having such a control could allow the MDS (or IR) system to choose either the more specific (reverse entailing) or general (forward entailing) sentence (or query) depending on the purpose of the summary (or user needs). 

To address the lack of entailment relation control in paraphrasing systems, we introduce a new task of \textbf{entailment relation aware paraphrase generation}: given a sequence and an entailment relation, generate a paraphrase which conforms to the given entailment relation. We consider three entailment relations (controls) in the spirit of monotonicity calculus~\citep{valencia1991studies}:
\begin{inparaenum}[(1)]
\item{\textbf{Equivalence ($\equiv$)} refers to semantically equivalent paraphrases ({\it e.g.} synonyms) where input sequence entails its paraphrase and vice-versa;}
\item{\textbf{Forward Entailment ($\sqsubset$)} refers to paraphrases that loose information from the input or generalizes it ({\it e.g.} hypernyms) {\it i.e.} input sequence entails its paraphrase;}
\item{\textbf{Reverse Entailment ($\sqsupset$)} refers to paraphrases that add information to the input or makes it specific ({\it e.g.} hyponyms) {\it i.e.} input sequence is entailed by its paraphrase.}
\end{inparaenum}

The unavailability of paraphrase pairs annotated with such a relation makes it infeasible to directly train a sequence-to-sequence model for this task.
Collecting such annotations for existing large paraphrase corpora such as ParaBank~\citep{hu2019parabank} or ParaNMT~\cite{wieting2018paranmt} is expensive due to scale.
We address this challenge in $3$ ways: \begin{inparaenum}[(1)] 
\item{by building a novel \textit{entailment relation oracle} based on natural language inference task (NLI)~\citep{bowman2015large,williams2018broad} to obtain weak-supervision for entailment relation for existing paraphrase corpora;}
\item{by recasting an existing NLI dataset, SICK~\citep{marelli2014sick}, into a small supervised dataset for this task, and}
\item{by proposing \textbf{E}ntailment \textbf{R}elation \textbf{A}ware \textbf{P}araphraser (\textbf{\texttt{ERAP}}) which is a reinforcement learning based (RL-based) weakly-supervised system that can be trained only using existing paraphrase and NLI corpora, with or without weak-supervision for entailment relation.}
\end{inparaenum}

Intrinsic and extrinsic evaluations show advantage of entailment relation aware (henceforth ``entailment-aware'') paraphrasing systems over entailment-unaware (standard uncontrolled paraphrase generation) counterparts.
Intrinsic evaluation of \texttt{ERAP} (via a combination of automatic and human measures) on recasted SICK (\S\ref{sec:recast}) dataset
shows that generated paraphrases conform to the given entailment relation with high accuracy while maintaining good or improved paraphrase quality when compared against entailment-unaware baselines. Extrinsic data-augmentation experiments (\S\ref{sec:data-augmentation}) on textual entailment task show that augmenting training sets using entailment-aware paraphrasing system leads to improved performance over entailment-unaware paraphrasing system, and makes it less susceptible to making incorrect predictions on adversarial examples.


\section{Entailment Relation Aware Paraphraser}
\subsubsection{Task Definition}
Given a sequence of tokens $X=[x_1,\dots,x_n]$, and an entailment relation $\mathscr{R}\in$\{Equivalence ($\equiv$), Forward Entailment ($\sqsubset$), Reverse Entailment ($\sqsupset$)\}, we generate a paraphrase $Y=[y_1,\dots,y_m]$ such that the entailment relationship between $X$ and $Y$ is $\mathscr{R}$. $\hat{Y}$ is the generated paraphrase and ${Y}$ is the reference paraphrase.

\begin{figure}
    \centering
    \scriptsize
    \includegraphics[width=0.40\textwidth,height=4.0cm]{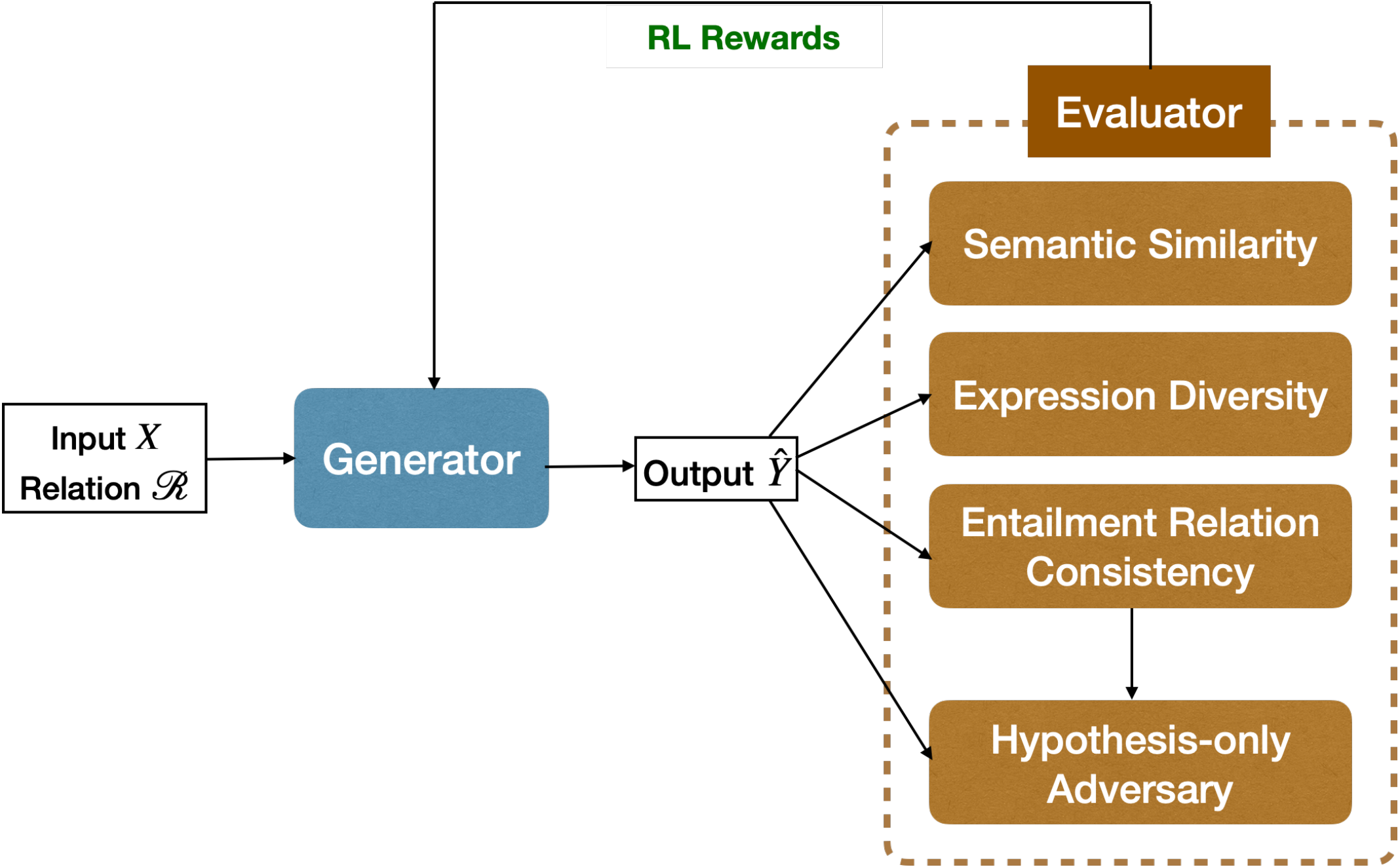}
    \caption{\textbf{\texttt{ERAP}}: Generator takes in a sequence $X$, an entailment relation $\mathscr{R}$, and outputs a paraphrase $\hat{Y}$. $\hat{Y}$ is scored by various scorers in the evaluator and a combined score (known as reward) is sent back to train the generator. Hypothesis-only adversary is adversarially trained on $\hat{Y}$ and predictions from the entailment relation consistency scorer.}
    \label{fig:framework}
\end{figure}

Neural paraphrasing systems~\citep{prakash2016neural,li2018paraphrase} employ a supervised sequence-to-sequence model to generate paraphrases. However, building a supervised model for this task requires paraphrase pairs with entailment relation annotations. To address this, we propose an RL-based paraphrasing system \texttt{ERAP} which can be trained with existing paraphrase and NLI corpora without any additional annotations. \textbf{\texttt{ERAP}} (Figure~\ref{fig:framework}) consists of a paraphrase generator 
(\S\ref{sec:generator}) and an evaluator (\S\ref{sec:evaluator}) comprising of various scorers to assess the quality of generated paraphrases for different aspects. Scores from the evaluator are combined (\S\ref{sec:rewards}) to provide feedback to the generator in the form of rewards. Employing RL allows us to explicitly optimize the generator over measures accounting for the quality of generated paraphrases, including the non-differentiable ones.

\subsection{Paraphrase Generator} \label{sec:generator}
The generator is a transformer-based~\citep{vaswani2017attention}  sequence-to-sequence model which takes ($X,\mathscr{R}$) and generates $\hat{Y}$. We denote the generator as $G$($\hat{Y}|X$,$\mathscr{R}$; $\theta_g$), where $\theta_g$ refers to parameters of the generator. We incorporate the entailment relation as a special token prepended to the input sequence. This way, entailment relation receives special treatment~\citep{kobus2017domain} and the generator learns to generate paraphrases for a given $X$, and $\mathscr{R}$.

\subsection{Paraphrase Evaluator} \label{sec:evaluator}
The evaluator comprises of several scorers to asses the quality of the generated paraphrase for three aspects: \textit{semantic similarity} with the input, 
\textit{expression diversity} from the input, and \textit{entailment relation consistency}. It provides rewards to the paraphrases generated by the generator as feedback which is used to update the parameters of the generator. We describe the various scorers below.
\subsubsection{Semantic Similarity Scorer }provides reward which encourages the generated paraphrase $\hat{Y}$ to have similar meaning as the input sequence $X$. We use MoverScore~\citep{zhao2019moverscore} to measure the semantic similarity between the generated paraphrase and the input, denoted as $r_s(X,\hat{Y})$. MoverScore combines contextualized representations with word mover's distance~\citep{kusner2015word} and has shown high correlation with human judgment of text quality. 
\subsubsection{Expression Diversity Scorer} rewards the generated paraphrase to ensure that it uses different tokens or surface form to express the input. We measure this aspect by computing n-grams dissimilarity (inverse BLUE~\citep{papineni2002bleu}),
\begin{equation} \label{eq:ib}
r_d(X,\hat{Y}) = 1-\text{BLEU}(\hat{Y},X)
\end{equation}
 Following~\citet{hu2019parabank}, we use modified BLEU without length penalty to avoid generating short paraphrases which can result in high inverse BLEU scores.
\subsubsection{Entailment Relation Consistency Scorer} is a novel scorer designed to reward the generated paraphrase in such a way that encourages it to adhere to the given entailment relation $\mathscr{R}$. To compute the reward, we build an oracle $O(X,Y)$ (details in \S\ref{sec:oracle}) based on natural language inferencing (NLI) and use likelihood of the given entailment relation from the Oracle as the score.
$
r_l(X,\hat{Y},\mathscr{R}) = O(l=\mathscr{R}|X,\hat{Y})
$.
As will be discussed further in \S\ref{sec:results}, we
found that entailment relation consistency scorer can result in generator learning simple heuristics ({\it e.g.} adding same adjective such as `desert', or trailing tokens like `and says' or `with mexico' for $\sqsupset$ or short outputs for $\sqsubset$) leading to degenerate paraphrases having high consistency score. 

Inspired by the idea of \textit{hypothesis-only} baselines~\citep{poliak2018hypothesis} for NLI task, we build a novel RoBERTa-based \textbf{\textit{Hypothesis-only} Adversary}, $A(l|\hat{Y})$, to penalize the generated paraphrases resorting to such heuristics. The adversary is a $3$-class classifier trained on the paraphrases generated during the training phase with the oracle prediction for ($X$,$\hat{Y}$) pair as the ground-truth. The Adversary loss $L(A)$ is
\begin{equation}
    L(A) = -\sum_{c=1}^{|C|}{O(X,\hat{Y})\log(A(l=c|\hat{Y}))},
\end{equation}
where $|C|=3$ is the number of entailment relations.
Training the adversary in this way helps in adapting to the heuristics taken by the generator during the course of training. The generator and the adversary are trained alternatively, similar to a GAN~\citep{goodfellow2014generative} setup. The penalty is computed as the likelihood of entailment relation being $\mathscr{R}$ using the adversary, 
$
 p_l(\hat{Y},\mathscr{R}) = A(l=\mathscr{R}|\hat{Y}).
$
We only penalize those generated paraphrases for which predicted relation is same as the input relation because incorrect prediction denotes no heuristic is taken by the generator. 

\subsection{Reinforcement Learning Setup}
The output paraphrases from the generator are sent to the scorers for evaluation. The various scores from the scorers are combined to give feedback (in the form of reward)
to the generator to update its parameters and
to improve the quality of the generated paraphrases conforming to the given relation. We emphasize that although the scores from our scorers are not differentiable with respect to $\theta_g$, we can still use them by employing RL (the REINFORCE algorithm) to
update the parameters of the generator~\citep{williams1992simple}. 

In RL paradigm, state at time $t$ is defined as $s_t=(X,\mathscr{R},\hat{Y}_{1:t-1})$ where $\hat{Y}_{1:t-1}$ refers to the first $t-1$ tokens that are already generated in the paraphrase. The action at time $t$ is the $t^{th}$ token to be generated. Let $V$ be the vocabulary, and $T$ be the maximum output length. The total expected reward of the current generator is then given by
$
J(G) = \sum_{t=1}^{T}{
\mathbb{E}_{\hat{Y}_{1:t-1}\sim G}
{[\sum_{y_t \in V}{\mathbb{P}(y_t|s_t)Q(s_t,y_t)}]
}
},
$
where $\mathbb{P}(y_t|s_t)$ is the likelihood of token $y_t$ given
the current state $s_t$, and $Q(y_t,s_t)$ is the cumulative discounted
reward for a paraphrase extended from $\hat{Y}_{1:t-1}$. The total reward, $Q$, is defined as the sum of the token level rewards.
\begin{equation}\label{eq:q}
Q(s_t,y_t) = \sum_{\tau=t}^{T}{\gamma^{\tau-t}r(s_\tau,y_\tau)},
\end{equation}
where $r(s_\tau, y_\tau)$ is the reward of token $y_\tau$ at state $s_\tau$, and $\gamma \in (0, 1)$ is a discounting factor so
that the future rewards have decreasing weights, since their estimates are less accurate. If we consider that $\hat{Y}_{1:t-1}$ has been given then for every $y_t$, the total expected reward becomes 
\begin{equation} \label{eq:j}
J(G) = \sum_{t=1}^{T}{
{\sum_{y_t \in V}{\mathbb{P}(y_t|s_t)Q(s_t,y_t)}
}}.
\end{equation}
\subsubsection{Sequence Sampling} 
To obtain $r(s_t, y_t)$ at each time step $t$, we need scores for each token. However, by design these scorers only evaluate complete sequences instead of single token or partial sequences. We therefore use the technique of rolling out~\citep{yu2017seqgan}, where the generator ``rolls out'' a given sub-sequence $\hat{Y}_{1:t}$ to generate complete sequence by sampling the remaining part of the sequence $\hat{Y}_{t+1:T}$. Following~\citet{gong2019reinforcement}, we use a combination of beam search and multinomial sampling to balance reward estimation accuracy at each time step and diversity of the generated sequence. We first generate a reference paraphrase $\hat{Y}^{ref}_{1:T}$ using beam search and draw $n$ samples of complete sequences $\hat{Y}_{1:T}$ by rolling out the sub-sequence $\hat{Y}^{ref}_{1:t}$ using multinomial sampling to estimate reward at each time step $t$.

\subsubsection{Reward Estimation} \label{sec:rewards}
We send $n$ samples of complete sequences drawn from the sub-sequence $\hat{Y}^{ref}_{1:t}$ to the scorers. The combined score $f(s_t, y_t)$ for an action $y_t$ at state $s_t$ is computed by averaging the score of the complete sequences rolled out from $\hat{Y}^{ref}_{1:t}$ defined as
\begin{equation}
\begin{split}
    f(s_t,y_t) = \frac{1}{n}\sum_{i=1}^{n}\alpha \cdot (r_l(X,\hat{Y}^i, \mathscr{R}) - p_l(\hat{Y}^i, \mathscr{R}))+\\
    \beta \cdot r_s(X,\hat{Y}^i) + \delta \cdot r_d(X,\hat{Y}^i),
\end{split}
\end{equation}
where $\alpha$, $\beta$, $\delta$, and $n$ are hyperparameters empirically set to $0.4$, $0.4$, $0.2$, and $2$, respectively. These parameters control the trade-off between different aspects for this multi-objective task. Following~\citet{siddique2020unsupervised}, we threshold\footnote{If $ 0.3\leq r_s(X,\hat{Y})\leq 0.98$ then the score is used as is otherwise, $0$. Similarly, if $r_s(X,\hat{Y})>0$ after thresholding then $r_d$, $r_l$, and $p_l$ are computed as defined, otherwise $0$.} the scorers' scores so that the final reward maintains a good balance across various scores. For example, generating diverse tokens at the expense of losing too much on the semantic similarity is not desirable. Similarly, copying the input sequence as-is to the generation is clearly not a paraphrase
({\it i.e.}, $r_s(X,\hat{Y})=1$). We define reward $r(s_t, y_t)$ for action $y_t$ at state $s_t$ as
\begin{equation}
r(s_t, y_t ) = \begin{cases}
f(s_t, y_t ) - f(s_{t-1}, y_{t-1}), &\quad t > 1, \\
f(s_1, y_1), &\quad t = 1
     \end{cases}
\end{equation}
The discounted cumulative reward $Q(s_t, y_t)$ is then computed from the rewards $r(s_\tau, y_\tau)$ at each time step using Eq.~\ref{eq:q} and the total expected reward is derived using Eq.~\ref{eq:j}
The generator loss $L(G)$ is defined as $-J(G)$.
\subsection{Training Details}
\subsubsection{Pre-training}
Pre-training has been shown to be critical for RL to work in unsupervised settings~\citep{siddique2020unsupervised,gong2019reinforcement} therefore, we pre-train the generator on existing large paraphrase corpora {\it e.g.} ParaBank~\citep{hu2019parabank} or ParaNMT~\cite{wieting2018paranmt} in two ways;
\begin{inparaenum}[(1)] 
\item{\textbf{Entailment-aware }uses Oracle (\S\ref{sec:oracle}) to obtain entailment relation for paraphrase pairs in the train set of paraphrase corpora, filter the semantically-divergent (\S\ref{sec:oracle}) pairs, upsample or downsample to have balanced data across relations, and train the generator with weak-supervision for entailment relation and gold-paraphrases, and} 
\item{\textbf{Entailment-unaware }trains the generator on paraphrase pairs as-is without any entailment relation.}
\end{inparaenum}
Pre-training is done in a supervised manner with the cross-entropy loss and offers immediate benefits for generator to learn paraphrasing transformations and have warm-start leading to faster model training.

\subsubsection{RL-based Fine-tuning}
We fine-tune the generator using feedback from the evaluator on recasted SICK
dataset (details in \S\ref{sec:recast}).
For any practical purposes, our RL-fine-tuning approach only requires input sequences without any annotations for entailment relation or ground-truth paraphrases. However, for a fair comparison against supervised or weakly-supervised baselines (\S\ref{sec:baselines}), we use the gold-entailment relation for recasted SICK 
during RL fine-tuning.

\section{Collecting Labeled Paraphrase Data}
Entailment-aware paraphrasing requires paraphrase pairs annotated with entailment relation. 
However, collecting entailment relation annotations for large size of  paraphrase corpora such as ParaBank\footnote{It consists of $~50$ million high quality English paraphrases obtained training a Czech-English neural machine translation (NMT) system and adding lexical-constraints to NMT decoding procedure.}~\citep{hu2019parabank} is too costly. 
To obtain entailment relations automatically for available paraphrase pairs, we train a NLI classifier and use it to derive the entailment relations as described below.

\noindent \textbf{Entailment Relation Oracle} \label{sec:oracle}
NLI is a standard natural language understanding task of determining whether a \textit{hypothesis} $h$ is true (entailment\footnote{Entailment in NLI is a uni-directional relation while Equivalence is a bi-directional entailment relation.} E), false (contradiction C), or undetermined (neutral N) given a \textit{premise} $p$~\citep{maccartney2009natural}. To build an entailment relation oracle, $O(X,Y)$, we first train a RoBERTa-based~\citep{liu2019roberta} $3$-class classifier, $o(l|\langle p,h\rangle)$, to predict the uni-directional (E, N, C) labels given a $\langle p,h \rangle$ pair. This classifier is then run forwards ($\langle X,Y \rangle$) and backwards ($\langle Y,X \rangle$) on the paraphrase pairs to get the uni-directional predictions which are further used to derive the entailment relations as follows.
\[
 \scriptsize
 \text{$O(X,Y)$} = \begin{cases}
       \equiv &\hspace{0.3pt} \text{if $o(l|\langle X,Y\rangle)=E$ \& $o(l|\langle Y,X\rangle)=E$ }\\
       \sqsubset &\hspace{0.3pt} \text{if $o(l|\langle X,Y\rangle)=E$ \& $o(l|\langle Y,X\rangle)=N$}\\
       \sqsupset & \hspace{0.3pt} \text{if $o(l|\langle X,Y\rangle)=N$ \& $o(l|\langle Y,X\rangle)=E$}\\
       \text{C} & \hspace{0.3pt} \text{if $o(l|\langle X,Y\rangle)=C$ \& $o(l|\langle Y,X\rangle)=C$} \\
       \text{N} &\hspace{0.3pt} \text{if $o(l|\langle X,Y\rangle)=N$ \& $o(l|\langle Y,X\rangle)=N$} \\
       \text{Invalid} &\hspace{0.3pt} \text{,otherwise}
     \end{cases}\]
The Oracle serves two purposes:
\begin{inparaenum}[(1)] 
\item{generates weak-supervision for entailment relations for existing paraphrase corpora, and} 
\item{asses the generated paraphrases for relation consistency.}
\end{inparaenum}
We only focus on $\equiv$, $\sqsubset$, and $\sqsupset$ relations as contradictory, neutral or invalid pairs are considered as semantically-divergent sentence pairs. 
\\
\noindent \textbf{Recasting SICK Dataset} \label{sec:recast}
SICK~\citep{marelli2014sick} is a NLI dataset created from 
sentences describing the same picture or video which are near paraphrases. 
It consists of sentence pairs $(p,h)$ with human-annotated NLI labels for both directions $\langle p,h\rangle$ and $\langle h,p\rangle$. We recast this dataset to obtain paraphrase pairs with entailment relation annotations derived using the gold bi-directional labels in the same way as $O$. We only consider the sentence pairs which were created by combining meaning-preserving transformations (details in~\ref{sec:app-sick}). We augment this data by adding valid samples obtained by reversing sentence pairs ($\forall$ p $\sqsubset$ h, we add h $\sqsupset$ p and $\forall$ p $\equiv$ h, we add h $\equiv$ p). Data statistics in Table~\ref{tab:datastat}.

\noindent \textbf{Oracle Evaluation}
We train the NLI classifier $o$ on existing NLI datasets namely, {MNLI}~\citep{williams2018broad}, {SNLI}~\citep{bowman2015large}, {SICK}~\cite{marelli2014sick} as well as diagnostic datasets such as, {HANS}~\citep{mccoy2019right}, others introduced in~\citet{glockner2018breaking,min2020syntactic}, using cross-entropy loss. Combining diagnostic datasets during training has shown to improve robustness of NLI systems which can resort to simple lexical or syntactic heuristics~\citep{glockner2018breaking,poliak2018hypothesis} to perform well on the task. 
The accuracy of $o(l|\langle p,h\rangle)$ on the combined test sets of the datasets used for training is $92.32\%$ and the accuracy of Entailment Relation Oracle $O(X,Y)$ on the test set of recasted SICK dataset is $81.55\%$. Before using the Oracle to obtain weak-supervision for entailment relation for training purposes, we validate it by manually annotating $50$ random samples from ParaBank. 
$78\%$ of the annotated relations were same as the Oracle predictions when C, N, and Invalid labels were combined.

\begin{table}[!t]
\centering
\scriptsize
\resizebox{0.80\columnwidth}{!}
{\begin{tabular}{l|cccc|ccc}
& \multicolumn{4}{c|}{\textbf{Recasted SICK}} & \multicolumn{3}{c}{\textbf{SICK NLI}}\\
\textbf{Split} & {$\equiv$}& {$\sqsubset$} & {$\sqsupset$} & {\textbf{Others}} & \textbf{E} & \textbf{N} & \textbf{C} \\
\hline
Train & $1344$&$684$ &$684$ & $420$ & $1274$ & $2524$& $641$\\
Dev & $196$&$63$&$63$ & $43$& $143$ & $281$ & $71$\\
Test & $1386$&$814$&$814$ & $494$& $1404$& $2790$& $712$\\
\hline
\end{tabular}}
\caption{E, N, C denote entailment, neutral, and contradiction, respectively. Others refers to neutral or invalid relation. }
\label{tab:datastat} 
\end{table}

\section{Intrinsic Evaluation}
Here we provide details on the entailment-aware and unaware comparison models, and the evaluation measures. 
\subsection{Comparison Models} \label{sec:baselines}
To contextualize \texttt{ERAP}'s performance, we train several related models including supervised and weakly-supervised, entailment-aware and unaware models to obtain lower and upper bound performance on recasted SICK as follows:
\begin{inparaenum}[(1)]
    \item{the generator is trained on recasted SICK in an entailment-aware (\textbf{Seq2seq-A}) and unaware (\textbf{Seq2seq-U}) supervised setting;}
    \item{the generator is pre-trained on ParaBank dataset in entailment-aware (\textbf{Pre-trained-A}) and unaware (\textbf{Pre-trained-U}) setting to directly test on the test set of recasted SICK;}
    \item{the pre-trained generators are fine-tuned on recasted SICK in entailment-aware (\textbf{Fine-tuned-A}) and unaware (\textbf{Fine-tuned-U}) supervised setting;}
    \item{multiple outputs ($k$$\in$$\{1,5,10,20\}$) are sampled using nucleus sampling~\citep{holtzman2019curious} from Seq2seq-U (\textbf{Re-rank-s2s-U}) or Fine-tuned-U (\textbf{Re-rank-FT-U}) and re-ranked based on the combined score $f(s_t,y_t)$. The highest scoring output is considered as the final output for {Re-rank-s2s-U} and {Re-rank-FT-U}.}
\end{inparaenum}
\subsection{Evaluation Measures}
\noindent \textbf{Automatic evaluation} to evaluate the quality of paraphrases is
primarily done using $i$BLEU~\citep{sun2012joint} which penalizes for copying from the input. Following~\cite{li2019decomposable,liu2020unsupervised} we also report BLEU~\citep{papineni2002bleu} (up to $4$ n-grams using sacrebleu library) and
Diversity (measured same as Eq.~\ref{eq:ib}) scores to understand the trade-off between these measures.
 We also compute, $\mathscr{R}$-Consistency, defined as the percentage of test examples for which the entailment relation predicted using oracle is same as the given entailment relation. \\
\noindent \textbf{Human evaluation} is conducted on $4$ aspects:
\begin{inparaenum}[(1)]
    \item \textbf{semantic similarity} which measures the closeness in meaning between paraphrase and input on a scale of $5$~\citep{li2018paraphrase};
    \item \textbf{diversity in expression} which measures if different tokens or surface-forms are used in the paraphrase with respect to the input on a scale of $5$~\citep{siddique2020unsupervised};
    \item \textbf{grammaticality} which measures if paraphrase is well-formed and comprehensible on a scale of $5$~\citep{li2018paraphrase};
    \item \textbf{relation consistency} which measures the \% of examples for which the annotated entailment relation is same as the input relation.
\end{inparaenum}
Three annotations per sample are collected for similarity, diversity, and grammaticality using Amazon Mechanical Turk (AMT), and the authors (blinded to the identity of the model and following proper guidelines) manually annotate for relation consistency as it is more technical and AMT annotators were unable to get the qualification questions correct. More details in Appendix~\ref{sec:app-human}. 
\begin{table}[!t]
\centering
\scriptsize
\resizebox{0.80\columnwidth}{!}
{\begin{tabular}{c|ccc|c}
{\textbf{Aware}} &{\textbf{BLEU$\uparrow$}} &
{\textbf{Diversity$\uparrow$}} &
{\textbf{$i$BLEU$\uparrow$}}
 & {\textbf{
$\mathscr{R}$-Consistency$\uparrow$}}\\
\hline
\xmark &$32.54$&$46.57$&$17.78$&$-$ \\
 \cmark &$\mathbf{33.08}$&$\mathbf{58.24}$&$\mathbf{19.06}$&$\mathbf{72.34}$ \\
\hline
\end{tabular}}
\caption{Evaluation of the generator pre-trained on ParaBank using entailment-aware (\cmark) and unaware (\xmark) settings.}
\label{tab:pretrain} 
\end{table}
\begin{table}[!t]
\centering
\scriptsize
\resizebox{0.99\columnwidth}{!}
{\begin{tabular}{l|c|ccc|c}
\toprule
\textbf{Model} & {$\mathscr{R}$-\textbf{Test}} &{\textbf{BLEU$\uparrow$}} &  {\textbf{Diversity$\uparrow$}} &
{\textbf{$i$BLEU$\uparrow$}}&
{$\mathscr{R}$-\textbf{
Consistency$\uparrow$}}\\
\hline
\rowcolor{lGray!200}
Pre-trained-U & \xmark  &$14.92$&$76.73$&$7.53$&$-$\\ 
\rowcolor{lGray!200}
Pre-trained-A & \cmark  &$17.20$&$74.25$&$\mathbf{8.75}$&$65.53$ \\
\hline
Seq2seq-U & \xmark  &$30.93$&$59.88$&$17.62$&$-$ \\
Seq2seq-A & \cmark  &$31.44$&$63.90$&$\mathbf{18.77}$&$38.42$ \\
\hline
Re-rank-s2s-U & \multirow{2}{*}{\cmark}  &$30.06$&$64.51$&$17.26$&$51.86$ \\
Re-rank-FT-U & &$41.44$&$53.67$&$\mathbf{23.96}$&$\mathbf{66.85}$ \\
\hline
\textbf{\texttt{ERAP}-U$^\star$} & \multirow{2}{*}{\cmark}  &$19.37$&$69.70$&$9.43$&$66.89$ \\
\textbf{\texttt{ERAP}-A} &  &$28.20$&$59.35$&$\mathbf{14.43}$&$\mathbf{68.61}$ \\
\hline
\rowcolor{Gray!150}
Fine-tuned-U & \xmark  &$41.62$&$51.42$ &$23.79$&$-$\\
\rowcolor{Gray!150}
Fine-tuned-A & \cmark  &$45.21$&$51.60$&$\mathbf{26.73}^\ast$&$\mathbf{70.24}^\ast$ \\
\hline
\hline
Copy-input & $-$ &$51.42$&$0.00$&$21.14$&$45.98$ \\
\bottomrule
\end{tabular}}
\caption{Automatic evaluation of paraphrases from \texttt{ERAP} against entailment-aware (A) and unaware (U) models described in \S\ref{sec:baselines}. $\mathscr{R}$-Consistency is measured only for models conditioned ($\mathscr{R}$-Test) on $\mathscr{R}$ at test time. Shaded rows denote \colorbox{Gray!150}{upper-} and \colorbox{lGray!200}{lower-}bound models.  
$\star$ denotes that only pre-training is done in entailment-unaware setting. \textbf{Bold-face} denotes best in each block and $\ast$ denotes best overall.}
\label{tab:autoeval-parabank} 
\end{table}
\subsection{Results and Analysis} \label{sec:results}
To use paraphrasing models for downstream tasks, we need to ensure that the generated paraphrases conform to the specified entailment relation and are of good quality.

\subsubsection{Automatic evaluation}
We first evaluate the pre-trained generators on a held-out set from ParaBank containing $500$ examples for each relation. Table~\ref{tab:pretrain} shows that the entailment-aware generator outperforms its unaware counterpart across all the measures. This boost is observed with weak-supervision for entailment relation demonstrating the good quality of weak-supervision. 

Next, We evaluate \texttt{ERAP} variants against the comparison models (\S\ref{sec:baselines}) on the recasted SICK test samples belonging to $\equiv,\sqsubset,\sqsupset$ relation and report the results in Table~\ref{tab:autoeval-parabank}\footnote{We report analogous results for ParaNMT in Appendix~\ref{sec:app-paranmt}.}.
\textit{Entailment-aware (-A) variants outperform corresponding unaware (-U) variants} on $i$BLEU score, while outperforming the majority-class ({\it i.e.}, $\equiv$) copy-input baseline (except for Seq2seq-A).
\textit{Weakly-supervised pre-training helps in boosting} the performance in terms of $i$BLEU and $\mathscr{R}$-Consistency score as evident from higher scores for Fine-tuned-A(U) model over Seq2seq-A(U). Poor performance of Seq2seq variants is because of the small dataset size and much harder multi-objective task. Re-ranking outputs from Fine-tuned-U achieve higher $i$BLEU and consistency score than Pre-trained-A which is explicitly trained with weak-supervision for relation. However, this comes at the computational cost of sampling multiple\footnote{We report results for $k\in\{1,5,10\}$ in the Appendix~\ref{sec:app-rerank}.} ($k$=$20$) outputs. Improved performance of Fine-tuned models over Seq2seq indicates the importance of pre-training. Both the \texttt{ERAP} variants achieve higher $i$BLEU and consistency than its lower-bounding (Pre-trained) models but the outputs show less diversity in expression and  make conservative lexical or syntactic changes. These results look encouraging until we notice Copy-input (last row) which achieves high BLEU and $i$BLEU, indicating that these metrics fail to punish against copying through the input (an observation consistent with~\citet{niu2020unsupervised}).

\begin{table}[!t]
\centering
\scriptsize
\resizebox{0.85\columnwidth}{!}
{\begin{tabular}{l|ccccc}
{\textbf{Model}}   & {\textbf{BLEU$\uparrow$}} & 
{\textbf{Diversity$\uparrow$}} &{\textbf{$i$BLEU$\uparrow$}} & {$\mathscr{R}$-\textbf{Consistency$\uparrow$}}\\
\hline
Gold-reference & $-$ & $48.58$&$-$ &  $81.55$\\
\hline
Pre-trained-A & $17.20$&$74.25$&$8.75$&$65.53$\\
\hline
+Con & $24.82$&$58.55$&$12.29$&$\underline{96.75}$\\
+Con+Sim &$39.78$& $\underline{42.05}$&$20.24$&$\underline{94.72}$\\
+Con+Sim+Div &$21.68$ &$68.41$&$11.29$&$\underline{93.60}$\\
\hline
\textbf{\texttt{ERAP-A}} & $28.20$&$40.65$&$14.43$&$68.61$\\
\hline
\end{tabular}}
\caption{Ablation of scorers in \texttt{ERAP}. Con, Sim, Div refers to relation consistency, semantic similarity, and expression diversity scorers. \underline{Underline} denote more copying of input for Diversity score and presence of heuristics in outputs for $\mathscr{R}$-Consistency score than gold-references.}
\label{tab:ablations} 
\end{table}
\subsubsection{Ablation analysis of each scorer}
We demonstrate the effectiveness of each scorer in \texttt{ERAP} via an ablation study in Table~\ref{tab:ablations}. Using only consistency scorer for rewarding the generated paraphrases, a significant improvement in consistency score is observed as compared to Pre-trained-A and Gold-references. However, this high score may occur at the cost of semantic similarity ({\it e.g.} $1$ in Table~\ref{tab:qualitative}) wherein output conforms to $\sqsubset$ relation at the cost of losing much of the content. Adding similarity scorer, helps in retaining some of the content (higher BLEU and $i$BLEU) but results in copying (low diversity) from the input. Addition of diversity scorer helps in introducing diversity in expression. However, model is still prone to heuristics ({\it e.g.} losing most of the content from input ($1$ in Table~\ref{tab:qualitative}), or adding irrelevant tokens `with mexico' or `desert' ($2$ in Table~\ref{tab:qualitative})) for ensuring high consistency score. Introducing Adversary reduces the heuristics learned by the generator. Together all the scorers help maintain a good balance for this multi-objective task.
\begin{table}[!th]
    \centering
    \scriptsize
  \resizebox{0.95\columnwidth}{!}
   { \begin{tabular}{p{1.4cm}p{7.0cm}}
    \hline
    \multicolumn{2}{c}{\textbf{1. Qualitative outputs for ablation study}}\\
    \hline
    \textbf{Input} $\sqsubset$ &a shirtless man is escorting a horse that is pulling a carriage along a road \\
    \textbf{Reference} &a shirtless man is leading a horse that is pulling a carriage \\
    \hline
    Only Con &a shirtless man escorts it.\\
    %
    Con+Sim&a shirtless man is escorting a horse.\\
  Con+Sim+Div&a shirtless man escorts a horse. \\
 \textbf{\texttt{ERAP-A}}&a shirtless person is escorting a horse who is dragging a carriage. \\
	\hline
	\hline
    \multicolumn{2}{c}{\textbf{2. Example of heuristic learned w/o Hypothesis-only Adversary}}\\
    \hline
    \textbf{Input} $\sqsupset$&a man and a woman are walking through a wooded area\\
    \textbf{Reference}&a man and a woman are walking together through the woods\\
    \hline
    -Adversary & a \textcolor{red}{desert} man and a woman walk through a wooded area\\
    \textbf{\texttt{ERAP}-A}& a man and a woman are walking down a path through a wooded area\\
    \hline
    \textbf{Input} $\sqsupset$& four girls are doing backbends and playing outdoors\\
    \textbf{Reference} &four kids are doing backbends in the park\\
    \hline
    -Adversary &four girls do backbends and play outside \textcolor{red}{with mexico}. \\
   \textbf{\texttt{ERAP}-A} &four girls do backbends and play games outside.\\
    \hline 
    \hline
    \multicolumn{2}{c}{\textbf{3. Output from comparison models and \texttt{ERAP}}}\\
    \hline
    \textbf{Input} $\equiv$& a ribbon is being twirled by a girl in pink \\
    \textbf{Reference}& a girl in pink is twirling a ribbon\\
        \hline
    Seq2seq-U & a girl is talking to a girl in pink \\
    Seq2seq-A & a girl in blue is surfing on a pink \\        
    Fine-tuned-U& a girl in pink is turning a tape\\
    Fine-tuned-A& a girl in pink is circling the ribbon.\\
    Re-rank-U&a girl is talking to a girl in a pink built \\
    Re-rank-A& a girl in a pink jacket is swinging a ribbon\\
    Pre-trained-U& a girl in pink is circling the ribbon. \\
    Pre-trained-A& the ribbon is being twisted by a girl in pink\\
    \texttt{ERAP}-U & the ribbon is being twisted by the girl in the pink\\
    \textbf{\texttt{ERAP}-A}& a girl in pink is twirling a ribbon.\\        
    \hline   
    \end{tabular}}
    \caption{Qualitative outputs: $1$ showing the effectiveness of various scorers, $2$ showing heuristic learned in the absence of hypothesis-only adversary, and $3$ from various models.  }
    \label{tab:qualitative}
\end{table}

\begin{table}[!ht]
\centering
\scriptsize
\resizebox{0.85\columnwidth}{!}
{\begin{tabular}{l|c|cccc}
{\textbf{Model}}&{$\mathscr{R}$-\textbf{Test}}&{\textbf{Similarity$\uparrow$}} & {\textbf{Diversity$\uparrow$}}  & {\textbf{Grammar$\uparrow$}} & {$\mathscr{R}$-\textbf{Consistency$\uparrow$}}\\
\hline
Pre-trained-U & \xmark &$4.60$&$2.62$&$4.73$&$-$\\
Pre-trained-A & \cmark &$4.67$&$2.60$&$4.67$&$48.00$\\
\hline
Re-rank-s2s-U & \multirow{2}{*}{\cmark} &$2.72$&$3.15$&$3.46$&$24.00$\\
Re-rank-FT-U &&$3.05$&$2.89$&$4.27$&$28.00$\\
\hline
\textbf{\texttt{ERAP}-U} & \multirow{2}{*}{\cmark} &$3.98$&$2.85$&$4.10$&$40.00$\\
\textbf{\texttt{ERAP}-A} & &$3.95$&$2.68$&$4.42$&$64.00$\\
\hline
Fine-tuned-U & \xmark &$3.87$&$3.10$&$4.83$&$-$\\
Fine-tuned-A & \cmark &$3.80$&$3.04$&$4.68$&$48.00$\\
\hline
\end{tabular}}
\caption{Mean scores across $3$ annotators are reported for Similarity ($\alpha$=$0.65$), Diversity ($\alpha$=$0.55$), and Grammaticality ($\alpha$=$0.72$) and \% of correct specified relation for $\mathscr{R}$-Consistency ($\alpha$=$0.70$). Moderate to strong inter-rater reliability is observed with Krippendorff's $\alpha$. }
\label{tab:humaneval} 
\end{table}
\subsubsection{Human evaluation}
We report the human evaluation for $25$ test outputs each from $8$ models for $4$ measures in Table~\ref{tab:humaneval}. $\texttt{ERAP}$-A achieves the highest consistency while maintaining a good balance between similarity, diversity and grammaticality. Re-rank-s2s-U has the highest diversity which comes at the cost of semantic similarity and grammaticality ({\it e.g.} $3$ in Table~\ref{tab:qualitative}). A strikingly different observation is high similarity and low diversity of Pre-trained variants, reinforcing the issues with existing automatic measures.
\begin{table*}[!t]
\centering
\scriptsize
\resizebox{0.80\pdfpagewidth}{!}
{\begin{tabular}{l|ccc|cc|ccc|cccccccc}
\toprule
\multirow{2}{*}{\textbf{Original}} & \multicolumn{15}{c}{\textbf{Augmentations}}\\
\cline{2-16}
& \multicolumn{3}{c|}{$\equiv$} &
\multicolumn{2}{c|}{$\sqsupset$}&\multicolumn{3}{c|}{$\sqsubset$} & \multicolumn{7}{c}{Unknown}\\
\hline
$\langle p,h\rangle$&$\langle p',h\rangle$&$\langle p,h'\rangle$&$\langle p',h'\rangle$&$\langle pr,h\rangle$&$\langle pr,h'\rangle$&$\langle p,hf\rangle$&$\langle p',hf\rangle$&$\langle pr,hf\rangle$&$\langle pf,h'\rangle$&$\langle pf,hf\rangle$&$\langle p,hr\rangle$&
$\langle p',hr\rangle$&$\langle pr,hr\rangle$&$\langle pf,r\rangle$&$\langle pf,h\rangle$\\
$E/NE$&$E/NE$&$E/NE$&$E/NE$&$E/NE$&$E/NE$&$E/U$&$E/U$&$E/U$&$U/U$&$U/U$&$U/U$&$U/U$&$U/U$&$U/U$&$U/U$\\
\bottomrule
\end{tabular}}
\caption{Original refers to the original sentence pair and its label (E (NE) denotes entails (does not entail)). Remaining columns denote various augmentation pairs and corresponding labels according to the entailment composition rules defined in \citet{maccartney2009natural}. $p'(h'), pr(hr), pf(hf)$ denote equivalent, reverse entailing, and forward entailing paraphrase, respectively.}
\label{tab:labelpreservation} 
\end{table*}
\section{Extrinsic Evaluation} \label{sec:data-augmentation}
The intrinsic evaluations show that \texttt{ERAP} produces quality paraphrases while adhering to the specified entailment relation. Next, we examine the utility of entailment-aware paraphrasing models over unaware models for a downstream application, namely paraphrastic data augmentation for textual entailment task.
Given two sentences, a premise $p$ and a hypothesis $h$, the task of textual entailment is to determine if a human would infer $h$ is true from $p$.
Prior work has shown that paraphrastic augmentation of textual entailment datasets improve performance~\citep{hu2019improved}; however, these approaches make the simplifying assumption that entailment relations are preserved under paraphrase, which is not always the case (see Figure~\ref{fig:task} and $~30$\% of ParaBank pairs were found to be semantically-divergent using Oracle).
We use SICK NLI dataset for this task because we have a paraphrasing system trained on similar data distribution\footnote{Note that we retained the train, test, development sets of SICK NLI dataset in the recasted SICK dataset and therefore the paraphrasing models have only seen train set.}. 

We hypothesize that entailment-aware augmentations will result in fewer label violations, and thus overall improved performance on the textual entailment task.
Moreover, explicit control over the entailment relation allows for greater variety of augmentations that can be generated (an exhaustive list of label preserving augmentations based on entailment relation between a $p$ (or $h$) and its paraphrase is presented in Table~\ref{tab:labelpreservation}) with entailment-aware models.

\noindent \textbf{Paraphrastic Data Augmentation}
We generate paraphrases for all premises $p\in P$ and hypotheses $h\in H$ present in the train set of SICK NLI using entailment-aware and unaware models. We obtain augmentation data by combining all the paraphrases (generated using entailment-aware models) with original data and label them as per Table~\ref{tab:labelpreservation}. Augmentation paraphrases generated from entailment-unaware models are (na\"ively) assumed to hold the $\equiv$ relation.
 RoBERTa-based binary classifiers are trained on original dataset along with the paraphrastic augmentations
to predict whether $p$ entails $h$. 

\noindent \textbf{Susceptibility to Augmentation Artifacts}
If paraphrastic augmentations introduce noisy training examples with incorrectly projected labels, this could lead to, what we call \textit{augmentation artifacts} in downstream models.
We hypothesize that paraphrastically augmented textual entailment (henceforth, PATE) models trained on entailment-aware augmentations will be less susceptible to such artifacts than models trained with entailment-unaware augmentations. To test this, we generate augmentations for the test set of SICK NLI and manually annotate $1253$ augmented samples to obtain $218$ incorrectly labeled examples. We evaluate PATE models on these examples (referred to as \textit{adversarial} test examples).
\begin{table}[!th]
\centering
\scriptsize
\resizebox{0.85\columnwidth}{!}{\begin{tabular}{l|c|cc|c}
{\textbf{Data}} &  {$\mathscr{R}$-\textbf{Test}} & {\textbf{Original-Dev$\uparrow$}} & {\textbf{Original-Test$\uparrow$}} & {\textbf{Adversarial-Test$\uparrow$}}\\
\hline
\textbf{SICK NLI} & -&$95.56$&$93.78$&$\mathbf{83.02}$\\
\hline
\textbf{+FT-U($\mathbf{\equiv}$)}
&\xmark&$95.15$&$93.68$&$69.72$\\
\hline
\textbf{+FT-A($\mathbf{\equiv}$)} &\multirow{4}{*}{\cmark}&$95.35$&$\mathbf{94.62}$&$77.98$\\
\textbf{+FT-A($\mathbf{\equiv,\sqsupset}$)} &&$\mathbf{95.76}$&$93.95$
&$75.69$\\
\textbf{+\texttt{ERAP}-A($\mathbf{\equiv}$)} &  &$95.15$&$94.58$&$78.44$\\
\textbf{+\texttt{ERAP}-A($\mathbf{\equiv,\sqsupset}$)} & &$95.15$&$93.86$&$69.72$\\
\hline
\end{tabular}}
\caption{Accuracy results on downstream data-augmentation experiments for textual entailment task. \textbf{FT/\texttt{ERAP}} refer to the Fine-tuned/proposed model used for generating augmentations. Type of augmentation used as per Table~\ref{tab:labelpreservation} in parenthesis. \textbf{U/A} denote entailment-unaware (aware) variant.}
\label{tab:adversarial} 
\end{table}

\noindent \textbf{Extrinsic Results}
We report accuracy of PATE models on original SICK development and test sets as well as on \textit{adversarial} test examples in Table~\ref{tab:adversarial}.
As per our hypothesis, models trained with augmentations generated using entailment-aware models result in improved accuracy on both original as well as adversarial test samples over those trained with entailment-unaware augmentations. Textual entailment model trained only on SICK NLI data performs the best on \textit{adversarial} test set as expected and proves that although augmentation helps in boosting the performance of a model, it introduces augmentation \textit{artifacts} during training.

\section{Related Work} \label{sec:relwork}
\textbf{Paraphrase generation} is a common NLP task with widespread applications. Earlier approaches are rule-based~\citep{barzilay1999information,ellsworth2007mutaphrase} or data-driven~\citep{madnani2010generating}. Recent, supervised deep learning approaches use LSTMs~\citep{prakash2016neural}, VAEs~\citep{gupta2018deep}, pointer-generator networks~\citep{see2017get}, and transformer-based~\citep{li2019decomposable} sequence-to-sequence models. \citet{li2018paraphrase} use RL for supervised paraphrasing.

\noindent \textbf{Unsupervised paraphrasing} is a challenging and emerging NLP task with limited efforts. \citet{bowman2015generating} train VAE to sample less controllable paraphrases. Others use metropolis-hastings~\citep{miao2019cgmh}, simulated annealing~\citep{liu2020unsupervised} or dynamic-blocking~\citep{niu2020unsupervised} to add constraints to the decoder at test time. \citet{siddique2020unsupervised} use RL to maximize expected reward based on adequacy, fluency and diversity. Our RL-based approach draws inspiration from this work by introducing oracle and hypothesis-only adversary.

\noindent \textbf{Controllable text generation} is a closely related field with efforts been made to add lexical~\citep{hu2019improved,garg2021unsupervised} or syntactic control~\citep{iyyer2018adversarial,chen2019controllable,goyal2020neural} to improve diversity of paraphrases. However, ours is the first work which introduces a semantic control for paraphrase generation. 

\noindent \textbf{Style transfer} is a related field that aims at transforming an input to adhere to a specified target attribute ({\it e.g.} sentiment, formality). 
RL has been used to explicitly reward the output to adhere to a target attribute~\citep{gong2019reinforcement,sancheti2020reinforced,luo2019dual,liu2020learning,goyal2021multi}. The target attributes are only a function of the output and defined at a lexical level. However, we consider a relation control which is a function of both the input and the output, and is defined at a semantic level. 

\section{Conclusion}
We introduce a new task of entailment-relation-aware paraphrase generation 
and propose a RL-based weakly-supervised model (\texttt{ERAP}) that can be trained 
without a task-specific corpus. Additionally, an existing NLI corpora is recasted to curate a small annotated dataset for this task, and provide performance bounds for it. A novel Oracle is proposed to obtain weak-supervision for relation control for existing paraphrase corpora. \texttt{ERAP} is shown to generate paraphrases conforming to the specified relation while maintaining quality of the paraphrase. 
Intrinsic and Extrinsic experiments demonstrate the utility of entailment-relation control, indicating a fruitful direction for future research. 

\bibliography{aaai22}

\clearpage
\appendix
\section{Implementation and Dataset Details} \label{sec:impl}
\subsection{Entailment Relation Oracle} \label{sec:app-oracle}
We use Ro{BERT}a-L architecture ($355$M parameters) from transformers library~\citep{wolf2019huggingface} to build NLI-classifier, $o(l|\langle p,h\rangle)$. We lower-case all the data and train the classifier for $3$ epochs ($\approx20$hrs) using cross-entropy loss. The model with best accuracy on the combined development splits of MNLI, SNLI, and SICK is used for final relation prediction. We use Adam optimizer with initial learning rate of $2e^{-5}$, warm up steps set at $0.06$ of total steps, batch size of $32$, and maximum input length of $128$. 

\subsection{Generator Pre-training} \label{sec:app-gen}
We build a transformers~\citep{vaswani2017attention} based encoder-decoder model with $6$ layers, $8$ attention heads, $512$ hidden state dimension, and $2048$ feed-forward hidden state size. The generator is trained for $20$ epochs on $4$GPUs and the model with best $i$BLEU score on the development set is used for testing in case of entailment-unaware setting. In the case of entailment-aware setting model with best score for harmonic mean of $i$BLEU and $\mathscr{R}$-Consistency on the development set is used for testing and further fine-tuning. We use Adam optimizer with initial learning rate of $1e^{-4}$, batch size of $256$, label smoothing and drop out rate $0.1$, maximum input length $40$, and minimum output length $5$. It took $2$ weeks to pre-train the generators. Entailment-unaware pre-training was done on ParaBank dataset which was partitioned into train ($93596425$) and validation ($1500$) split. We also run analogous experiments on ParaNMT partitioned into train ($5367128$), validation ($1000$), and test ($2000$) sets. For entailment-aware pre-training, we filter paraphrase pairs for which the oracle predicted entailment relation was anything other than $\equiv$, $\sqsubset$, or $\sqsupset$ and downsample the majority relation ($\equiv$) in ParaBank to get $21797111$ ($7570204$ belong to $\sqsubset$, $7366240$ to $\sqsupset$ , and rest $\equiv$) examples. For ParaNMT, we upsample the minority relations to get $9480295$ ($3330183$ belong to $\sqsubset$, $3233640$ to $\sqsubset$, and remaining $\equiv$) examples.

\subsection{Generator Fine-tuning} \label{sec:app-genfine}
For RL-finetuning, we use the same training setting as entailment-aware model mentioned above but use learning rate of $2e{-5}$, batch size of $16$, and fine-tune on $1$ GPU. We experiment with $\alpha={0.4,0.5,0.3}$, $\beta={0.4,0.5,0.3}$, and $\delta={0.2,0.3}$ and found $0.4,0.4,0.2$ respectively to give best results. The discounting factor $\gamma$ was set at $0.99$. Sequences were sampled using nucleus sampling~\citep{holtzman2019curious} probability of $0.8$, and temperature set at $1.0$. We filter examples belonging to entailment relation other than $\equiv$, $\sqsubset$, or $\sqsupset$, from recasted SICK and upsample the minority relation to fine-tune the pre-trained generator using RL.

\subsection{Evaluator} \label{sec:app-eval}
We use python's moverscore library \url{https://pypi.org/project/moverscore/} to compute moverscore for semantic similarity, sacrebleu \url{https://pypi.org/project/sacrebleu/} to compute BLEU, and RoBERTa-L classifier for the Adversary with same configuration as Entailment Relation Oracle. 

\subsection{Inference} \label{sec:app-infer}
We use beam search with beam width of $5$, minimum and maximum output length set at $5$ and $40$, respectively. 

All the models were trained and tested on $8$ NVIDIA Tesla V$100$ SXM$2$ $32$GB GPU machine. 

\section{Recasting SICK} \label{sec:app-sick}
We use examples $\in$\{$S1aS2a$, $S1aS2b$, $S1bS2a$, $S1bS2b$, $S1aS1b$\} meaning preserving transformations. In case of bi-directional entailing examples, we double the data by using both original as well as reverse pair and label it as $\equiv$. In case of entailment from only premise to hypothesis, we add this sample with $\sqsubset$ relation and reverse sample with $\sqsupset$ relation.

\section{Additional Results} \label{sec:app-add-results}
\subsection{Re-ranking Baseline Results} \label{sec:app-rerank}
Re-rank baselines are built over Seq2seq-U (Fine-tuned-U) resulting in Re-rank-s2s-U (Re-rank-FT-U).
Seq2seq-U (or Fine-tuned-U) is used to generate multiple outputs ($k$$\in$$\{1,5,10,20\}$) for an input using nucleus sampling~\citep{holtzman2019curious} and the outputs are scored with the scorers for a desired entailment relation to obtain a combined score $f(s_t,y_t)$ for each output. Highest scoring output is considered as the final output. Re-ranking results for $k\in\{1,5,10\}$ are reported in Table~\ref{tab:rerank-parabank}.
\begin{table}[!t]
\centering
\scriptsize
\resizebox{0.99\columnwidth}{!}
{\begin{tabular}{l|c|c|ccc|c}
\toprule
\textbf{Model} & {$\mathscr{R}$-\textbf{Test}}
&\textbf{k}
&{\textbf{BLEU$\uparrow$}} &  {\textbf{Diversity$\uparrow$}} &
{\textbf{$i$BLEU$\uparrow$}}&
{$\mathscr{R}$-\textbf{
Consistency$\uparrow$}}\\
\hline
Re-rank-s2s-U & \multirow{3}{*}{\cmark}  & $1$&$26.57$&$66.10$&$14.91$&$23.76$ \\
Re-rank-s2s-U &  & $5$&$29.08$&$65.27$&$16.60$&$39.68$ \\
Re-rank-s2s-U &  &$10$&$29.77$&$64.78$&$\mathbf{17.02}$&$\mathbf{46.45}$ \\
\hline
Re-rank-FT-U & \multirow{3}{*}{\cmark}& $1$&$38.49$&$45.16$&$22.03$&$40.15$ \\
Re-rank-FT-U & &$5$&$41.24$&$44.16$&$\mathbf{23.93}$&${56.67}$ \\
Re-rank-FT-U & &$10$&$41.30$&$53.88$&$23.91$&$\mathbf{62.67}$ \\
\bottomrule
\end{tabular}}
\caption{Additional Re-rank baseline results}
\label{tab:rerank-parabank} 
\end{table}
\begin{table}[!h]
\centering
\scriptsize
\resizebox{0.99\columnwidth}{!}
{\begin{tabular}{l|c|ccc|c}
\toprule
\textbf{Model} & {$\mathscr{R}$-\textbf{Test}} &{\textbf{BLEU$\uparrow$}} &  {\textbf{Diversity$\uparrow$}} &
{\textbf{$i$BLEU$\uparrow$}}&
{$\mathscr{R}$-\textbf{
Consistency$\uparrow$}}\\
\hline
\rowcolor{lGray}
Pre-trained-U & \xmark  &$17.15$&$71.05$&$8.23$&$-$\\ 
\rowcolor{lGray}
Pre-trained-A & \cmark  &$20.57$&$67.89$&$\mathbf{10.05}$&$\mathbf{66.52}$ \\
\hline
Seq2seq-U & \xmark  &$30.93$&$59.88$&$17.62$&$-$ \\
Seq2seq-A & \cmark  &$31.44$&$63.90$&$\mathbf{18.77}$&$\mathbf{38.42}$ \\
\hline
Re-rank-FT-U (k=1) & \multirow{4}{*}{\cmark}&$36.65$&$57.45$&$21.11$&$38.65$ \\
Re-rank-FT-U (k=5) & &$39.83$&$55.60$&$23.09$&$58.36$ \\
Re-rank-FT-U (k=10) & &$39.45$&$55.45$&$22.77$&$64.20$ \\
Re-rank-FT-U (k=20) & &$40.49$&$54.94$&$\mathbf{23.50}$&$\mathbf{69.81}$ \\
\hline
\textbf{\texttt{ERAP}-U$^\star$} & \multirow{2}{*}{\cmark}  &$22.04$&$63.71$&$10.37$&$63.07$ \\
\textbf{\texttt{ERAP}-A} &  &$28.10$&$54.21$&$\mathbf{13.32}$&$\mathbf{73.95}^\ast$ \\
\hline
\rowcolor{Gray}
Fine-tuned-U & \xmark  &$40.99$&$52.81$ &$23.70$&$-$\\
\rowcolor{Gray}
Fine-tuned-A & \cmark  &$43.75$&$53.59$&$\mathbf{26.08}^\ast$&$\mathbf{65.53}$ \\
\hline
\hline
Copy-input & $-$ &$51.42$&$0.00$&$21.14$&$45.98$ \\
\bottomrule
\end{tabular}}
\caption{Automatic evaluation of paraphrases from \texttt{ERAP} against entailment-aware (A) and unaware (U) models. $\mathscr{R}$-Consistency is measured only for models conditioned ($\mathscr{R}$-Test) on $\mathscr{R}$ at test time. Shaded rows denote \colorbox{Gray!150}{upper-} and \colorbox{lGray!200}{lower-}bound models.  
$\star$ denotes that only pre-training is done in entailment-unaware setting. \textbf{Bold-face} denote best in each block and $\ast$ denote best overall.}
\label{tab:autoeval-paranmt} 
\end{table}
\subsection{Analogous Results for ParaNMT} \label{sec:app-paranmt}
We pre-train the generator with $5$M version of  ParaNMT~\citep{wieting2018paranmt} dataset in entailment-aware and unaware settings and fine-tune on recasted SICK dataset. The results are shown in Table~\ref{tab:autoeval-paranmt}. Re-rank-s2s-U are same as ParaBank Table in the main paper.

\section{Automatic Evaluation}
BLEU score is computed against one gold-paraphrases for recasted SICK and multiple gold-references available for ParaBank dataset. 

\section{Human Evaluation Details} \label{sec:app-human}
Each annotator is asked to rate paraphrase pairs from $8$ models for an aspect and $4$ additional test question annotations are obtained for quality check. If all the test questions are answered correctly then only annotations are considered else discarded. 
The Liker scale for various aspects is defined as. 

\noindent \textbf{Semantic similarity }$1$-completely different meaning, $2$-different meaning, $3$-slightly similar in meaning, $4$-mostly similar in meaning, $5$-exactly same in meaning.

\noindent \textbf{Diversity in Expression }$1$-exactly same, $2$-minor changes (a$\rightarrow$the), $3$-slightly different, $4$-different, $5$-very different.

\noindent \textbf{Grammaticality }$1$-completely nonsensical, $2$-somewhat nonsensical, $3$-major grammatical errors, $4$-minor grammatical errors, $5$-no grammatical error. Only paraphrased sentence is annotated.

\noindent \textbf{Relation consistency} annotation is done by the authors because it is more technical. Each paraphrase pair is annotated with one of the following relations; equivalence, forward entailment, reverse entailment, contradiction, neutral, or none of these and majority relation for an example is compared against the input relation. The $\%$ of examples annotated same as the input relation is reported in human evaluation table. There were $11$, $8$, and $6$ samples belonging to equivalence, forward entailment, and reverse entailment relation, respectively out of the $25$ randomly sampled examples which were annotated.

\end{document}